\pdfoutput=1

\documentclass[11pt]{article}

\usepackage{EMNLP2022}

\usepackage{float}
\usepackage{times}
\usepackage{latexsym}
\usepackage{multirow}
\usepackage{graphicx}
\usepackage[T1]{fontenc}
\usepackage{booktabs}

\usepackage[utf8]{inputenc}

\usepackage{microtype}

\newcommand{\taskname}{\texttt{OpenStance}}
\newcommand{\masktopic}{\texttt{MASK-Topic}}
\newcommand{\masktext}{\texttt{MASK-Text}}

\title{\taskname: Real-world Zero-shot Stance Detection}

\graphicspath{ {./images/} }

\author{Hanzi Xu, Slobodan Vucetic \and Wenpeng Yin \\
  Temple University\\
  {\tt
    \{hanzi.xu; slobodan.vucetic; wenpeng.yin\}@temple.edu}}

\begin{document}
\maketitle


\begin{abstract}
Prior studies of zero-shot stance detection identify the attitude of texts towards unseen topics occurring in the same document corpus. Such task formulation has three limitations: (i) \textit{Single domain/dataset}. A system is optimized on a particular dataset from a single domain; therefore, the resulting system cannot work well on other datasets; (ii) the model is evaluated  on \textit{a limited number of unseen topics}; (iii) it is assumed that \textit{part of the topics has rich annotations}, which might be impossible in real-world applications.  These drawbacks will lead to an impractical stance detection system that fails to generalize to open domains and open-form topics.

This work  defines \taskname: open-domain zero-shot stance detection, aiming to handle stance detection in an open world with neither domain constraints nor topic-specific annotations. The key challenge of \taskname~lies in the \textit{open-domain generalization}: learning a system with fully unspecific supervision but capable of generalizing to any dataset. To solve \taskname, we propose to combine  \textbf{indirect supervision}, from textual entailment datasets, and \textbf{weak supervision}, from data generated automatically by pre-trained Language Models. Our single system, without any topic-specific supervision, outperforms the supervised method on three popular datasets.
To our knowledge, this is the first work that studies stance detection under the open-domain zero-shot setting. All data and code are publicly released.\footnote{\url{https://github.com/xhz0809/OpenStance}}
\end{abstract}

\section{Introduction}

Stance detection differentiates the attitude (e.g., \texttt{support}, \texttt{oppose}, or \texttt{neutral}) of a text towards a topic \cite{DBLPalkerAAG12}. The topic can be a  phrase   or a complete sentence. The same text can express the author's positions on many different topics. For example, a tweet on climate warming may also express attitudes about environmental policies as well as the debate between electric or fuel cars. Such compound expression can be seen on all online platforms, including News outlets, Twitter, blogs, etc. Therefore, stance detection can be a complicated task that is essential for developing the inference capability of NLP models as well as other disciplines such as politics, journalism, etc.

Since the textual expressions and  the size of topics in the real world are unpredictable, zero-shot stance detection has become the mainstream research direction in this area: topics in the test set are unseen during training. For example, \citet{mohammad2016semeval} created a dataset SemT6 based on tweets with six  noun phrases as topics. One of the topics was reserved for testing and the remaining were used for training. \citet{allaway2020zero} extended the topic size on the domain of news comments by covering 4,000 topics in training  and 600 unseen topics in testing.

However, despite the change in the domain and topic size, there are three major limitations in  previous  studies \textit{which make the task not a real zero-shot task}: (i) the dataset only contains texts from a single domain, such as news comments in VAST \cite{allaway2020zero} and tweets in SemT6 \cite{mohammad2016semeval}; (ii) most literature studied only a limited size of topics with a single textual form (either noun phrases or sentential claims), e.g., \cite{mohammad2016semeval,conforti2020will}; (iii) rich annotation for at least part of the topics is always required, which is not possible in real-world applications because data collection can be very time-consuming and costly \cite{enayati-etal-2021-visualization}. Those limitations lead to an impractical zero-shot stance detection system that cannot generalize well to unseen domains and open-form topics. 

In this work, we re-define what a zero-shot stance detection should be. Specifically, we define \taskname: an open-domain zero-shot stance detection, aiming to build a system that can work in the real world without any specific attention to the text domains or topic forms. More importantly, no task-specific supervision is needed. To achieve this, we propose to combine two types of supervision: \emph{indirect supervision} and \emph{weak supervision}. The indirect supervision comes from textual entailment---we treat the stance detection problem as a textual entailment task since the attitude toward a topic should be inferred from the input text. Therefore, the existing entailment datasets, such as MNLI \cite{DBLPWilliamsNB18}, can contribute supervision to the zero-shot setting. To collect supervision that is more specific to the \taskname~task, we design two MASK choices (\texttt{MASK-topic} and \texttt{MASK-text}) to prompt GPT-3 \cite{brown2020language} to generate weakly supervised data. Given an input text and a stance label (\texttt{support}, \texttt{oppose}, or \texttt{neutral}), \texttt{MASK-topic}  predicts what topic is appropriate based on the content; given a topic and a label, \texttt{MASK-text} seeks the text that most likely holds this stance. The collection of weakly supervised data only needs the unlabeled texts and the set of topics that users want to include. The joint power of indirect supervision and weak supervision will be evaluated on VAST, SemT6 and Perspectrum \cite{chen2019seeing}, three popular datasets that cover distinct domains, different sizes and diverse textual forms of topics. Experimental results show that although no task-specific supervision is used, our system can get robust performance on all three datasets, even outperforming the task-specific supervised models (72.6 vs. 69.3 by mean F1 over the three datasets).

Our contributions are threefold: (i) we define \taskname, an open-domain zero-shot stance detection task, that fulfills real-world requirements while having never been studied before; (ii) we design a novel masking mechanism to let GPT-3 generate weakly supervised data for \taskname. This mechanism can inspire other NLP tasks that detect relations between two pieces of texts; (iii) our approach, integrating indirect supervision and weak supervision, demonstrates outstanding generalization among three datasets that cover a wide range of text domains, topic sizes and topic forms.

\section{Related Work}

\paragraph{Stance detection.} Stance detection, as a recent member of the NLP family, was mainly driven by newly created datasets. In the past studies, datasets have been constructed from diverse domains like online debate forums \cite{walker-etal-2012-corpus,hasan-ng-2014-taking,abbott-etal-2016-internet}, news comments \cite{krejzl2017stance,lozhnikov2018stance}, Twitter \cite{mohammad2016semeval,kuccuk2017stance,tsakalidis2018nowcasting}), etc. 

\paragraph{Zero-shot stance detection.} 
Recently, researchers started to work on zero-shot stance detection in order to build a system that can handle unseen topics. Most work split the collected topic-aware annotations into \emph{train} and \emph{test} within the same domain. 
\citet{allaway2020zero} made use of topic similarity to connect unseen topics with seen topics. \citet{allaway2021adversarial} designed adversarial learning to learn domain-independent information and topic-invariant representations. Similarly, \citet{wang2021solving} applied adversarial learning to extract stance-related but domain-invariant features existed among different domains. \citet{liu2021enhancing} utilized common sense knowledge from ConceptNet \cite{speer2017conceptnet} to introduce extra knowledge of the relations between the texts and topics. Most prior systems worked on a single domain and were tested on a small number of unseen topics. \citet{li2021improving} tried to test on various unseen datasets by jointly optimizing on multiple training datasets. However, they still assumed that part of the topics or domains has rich annotations. In contrast, our goal is to design a system that can handle stance detection in an open world without requiring any domain constraints or topic-specific annotations.

\paragraph{Textual entailment as indirect supervision.} Textual entailment studies if a hypothesis can be entailed by a premise; this was proposed as a unified inference framework for a wide range of NLP problems \cite{dagan2005pascal}.  Recently, textual entailment is widely utilized to help solve many tasks, such as few-shot intent detection \cite{fstc}, ultra-fine entity typing \cite{lite}, coreference resolution \cite{yinentailment}, relation extraction \cite{fstc,DBLPSainzLLBA21}, event argument extraction \cite{DBLP01376}, etc. As far as we know, our work is the first one that successfully leverages the indirect supervision from textual entailment for stance detection.

\paragraph{Weak supervision from GPT-3.} As the currently most popular and (arguably) well-behaved pre-trained language model, GPT-3 \cite{brown2020language} has been a great success on few-shot and zero-shot NLP. As an implicit knowledge base fully in the form of parameters, it is not surprising that researchers attempt to extract knowledge from it to construct synthetic data, e.g., \cite{yoo2021gpt3mix,wang2021want}. We use GPT-3 to collect distantly supervised data by two novel masking mechanisms designed specifically for the \taskname.


\section{Problem definition}\label{sec:taskdefinition}
\taskname~has the following requirements: 
\begin{itemize}
    \item An instance includes three items: text $s$, topic $t$ and a stance label $l$ ($l\in\{\texttt{support}, \texttt{oppose}, \texttt{neutral}\}$); the task is to learn the function $f(s,t)\rightarrow l$;
    \item The text $s$ can come from any domain; the topic $t$ can be any textual expressions, such as a noun phrase ``gun control'' or a sentential claim ``climate change is a real concern'';
    \item All labeled instances \{($s$, $t$, $l$)\} only exist in \emph{test}; no \emph{train} or \emph{dev} is provided;
    \item Previous work used different metrics for the evaluation. For example, VAST \cite{allaway2020zero} used macro-averaged F1 regarding stance labels, while studies on SemT6 \cite{allaway2021adversarial,DBLPLiangC00YX22} reported the  F1 scores per topic. To make systems be comparable, we unify the evaluation and use the label-oriented macro F1 as our main metric. 
    
\end{itemize}

\paragraph{\taskname~ vs. prior zero-shot stance detection.} Prior studies of zero-shot stance detection worked on a single dataset $D^i$ in which all texts $s$ comes from the same domain. Topics $t$ in the dataset are  split into $train$, $dev$  and $test$ disjointly. The main issue is that a model that fits $D^i$ does not work well on a new dataset $D^j$ that may contain $s$ of different domains and unseen $t$. For example, a model trained on VAST can  only get F1 49.0\% on Perspectrum, which is around the performance of random guess.
\taskname~ aims at handling multiple datasets of open domains and open-form topics without looking at their $train$ and $dev$.

\paragraph{\taskname~ vs. textual entailment.} Stance detection is essentially a textual entailment problem if we treat the text $s$ as the premise, and the stance towards the topic $t$ as the hypothesis. This motivates us to use indirect supervision from textual entailment to deal with the stance detection problem. Nevertheless, there are two distinctions between them: (i) even though we can match $l$ of stance detection with the labels of textual entailment: \texttt{support} $\rightarrow$ \texttt{entailment}, \texttt{oppose} $\rightarrow$ \texttt{contradict} and \texttt{neutral} $\rightarrow$ \texttt{neutral}, whether a topic $t$ in stance detection can be treated as a hypothesis depends on the text form of $t$. If $t$ is noun phrases such as ``gun control'', $t$ cannot act as a hypothesis alone as there is no stance in it; if $t$ is a sentential claim such as ``climate change is a real concern'', inferring the truth value of this hypothesis  is exactly a textual entailment problem. This observation motivates us to test \taskname~ on topics of both phrase forms and sentence forms; (ii) Zero-shot textual entailment means the size of  the annotated instances for \emph{labels} is zero, while \taskname~requires the \emph{topics} have zero labeled examples. 

\section{Methodology}
This section introduces how we collect and combine \emph{indirect supervision} and \emph{weak supervision} to solve \taskname.

\paragraph{Indirect Supervision.} As we discussed in Section \ref{sec:taskdefinition}, stance detection is a case of textual entailment since the stance $l$ towards a topic $t$ should be inferred from the text $s$. To handle the zero-shot challenge in \taskname, textual entailment is a natural choice for indirect supervision. 

Specifically, we first cast stance detection instances into the textual entailment format by combining $l$ and $t$ as a sentential hypothesis $h$, such as ``it \texttt{supports} \texttt{topic}'', and treating the $s$ as the premise $p$; then a pretrained model on MNLI \cite{DBLPWilliamsNB18}, one of the largest entailment dataset, is ready to predict the relationship between the  $p$ and  $h$. An entailed (resp. contradicted or neutral) $h$ means the topic $t$ is supported (resp. opposed or neutral) by the text $s$.

Unfortunately, the indirect supervision from textual entailment may not perform well enough in real-world \taskname~considering the widely known brittleness of pretrained entailment models and the open domains and open-form topics in \taskname. Therefore, in addition to the indirect supervision from textual entailment, we will collect weak supervision that is aligned better with the texts $\{x\}$ and the  topics $\{t\}$. 

\paragraph{Weak Supervision.} 

For the next step, we would like to create some weakly supervised data using easily available resources to obtain a better understanding of the target task. We used GPT-3 \cite{brown2020language}, a pre-trained autoregressive language model that can perform text completion at (arguably) a near-human level, to help us create some weakly labeled instances. 

We form  incomplete sentences using prompts, and let the GPT-3 complete them. Since a stance label $l$ connects the text $s$ and the topic $t$ and such connection is unavailable in a zero-shot setting, the construction of incomplete sentences is driven by two questions: (i) given an input text $s$ and a stance, e.g., \texttt{support}, what topics are supported by $s$? (ii) given  a topic  and a stance, for example, \texttt{support},  what texts support this topic? As a result,  there are two kinds of prompts: \masktopic~and \masktext. To implement the two masking mechanisms, we need to prepare three sets: the raw texts \{$s$\}, a set of topics \{$t$\}, and the known stance labels \{\texttt{support}, \texttt{oppose}, \texttt{neutral}\}. It is noteworthy that no topic-specific human annotations are used here.

\textbullet \textbf{\masktopic}: In this masking framework, we randomly choose a \texttt{text} from \{$s$\} and a stance label from \{\texttt{support}, \texttt{oppose}, \texttt{neutral}\}, then build the prompt as:
\fbox{%
  \begin{minipage}{0.95\linewidth}
     S/he claims  \texttt{text}, so s/he  \texttt{label} the idea of \underline{\textsc{MASK}}
  \end{minipage}
}

For example, when  the \texttt{text} is “Coldest and wettest summer in memory” and the label is \texttt{oppose}, the prompt would be ``S/he claims coldest and wettest summer in memory, so s/he opposes the idea of”. Then, this prompt is fed into GPT-3, and the completion ``global warming'' would be the predicted topic.

\textbullet \textbf{\masktext}: In this case, we randomly choose a \texttt{topic} from \{$t$\} and a stance $\texttt{label}$ towards it, then build the prompt as:
\fbox{%
  \begin{minipage}{0.95\linewidth}
     His/her attitude towards $\texttt{topic}$ is $\texttt{label}$ because s/he thinks \underline{$\textsc{MASK}$}
  \end{minipage}
}

For example, when the \texttt{topic} is ``climate change is a real concern'', the \texttt{label} is ``oppose'', the completed sentence filled by GPT-3 could be ``His attitude towards climate change is a real concern is opposition because s/he thinks \underline{the science behind climate change is not settled}''.


For any  dataset of stance detection, we first collect the three sets (i.e., \{$s$\}, \{$t$\}, and \{$l$\}) \emph{from the label-free training set} without peeking at any gold annotations, then use  \masktopic~and \masktext~ prompts to generate equal number of weakly supervised examples. We will study which masking scheme is more effective in experiments. In addition, to have a fair comparison with supervised methods that learn on the $train$ of a task, we make sure   our generated weakly supervised data has the  same size as the $train$ for any target task.

Although noise is common in weakly supervised data, GPT-3 performs badly on \texttt{neutral} completions for both \masktopic~and \masktext~tasks. This is not a surprise for the \masktopic~since the GPT-3 is asked to provide a topic that the given text has a neutral attitude for, while most texts, obtained from unlabeled \emph{train} and originally extracted from social networks, usually express a strong attitude. Furthermore, in  \masktext, even though the GPT-3 can output a text given the \texttt{neutral} label towards a topic, the response is very general and does not provide any insights. For example, when the template is "His attitude towards high school writing skills is \texttt{neutral} because he thinks [MASK]", GPT-3 fills out the MASK with "that they are important but not essential." Obviously, it is much easier to generate text with a clear attitude compared to a neutral stance. On the one hand, GPT-3 may not really understand what a \texttt{neutral} stance is. On the other hand, even humans cannot easily write a neutral opinion towards a topic. Since the quality of generated \texttt{neutral} instances is not very promising, we take the same approach as how VAST \cite{allaway2020zero} collected its neutral samples: matching texts with random topics in the dataset.

\paragraph{Training strategy.} To keep consistent format and make full use of the entailment reasoning framework, we convert all phrase-form \texttt{topic} in the weak supervision data into a  sentence-form hypothesis with the positive stance, i.e., ``he is in favor of \texttt{topic}'' (note that this does not change the original label). Then, we randomly split the weak supervision data as \emph{train} (80\%) and \emph{dev} (20\%). Given the entailment dataset MNLI  as the indirect supervision data ($D_{ind}$) and weakly supervised data ($D_{weak}$) from GPT-3, we first pretrain a RoBERTa-large \cite{DBLP11692} on $D_{ind}$, then finetune on  $D_{weak}$. 
In inference, we test the final model on the \emph{test} of each task, checking the system's generalization ability on diverse domains without optimizing on any domain-specific \emph{train}.

\section{Experiments}

\begin{table}[t]
\setlength{\tabcolsep}{4pt}
  \centering
  \begin{tabular}{l|cccc}
  \toprule
  & \multirow{2}{*}{domain} & \#topic &
  topic   & \multirow{2}{*}{\#labels}  \\
  & & train/test & form &  \\
  \midrule\midrule
  SemT6 & tweet & 6& phrase& 3  \\
  VAST &  debate & 4641/600 & phrase&  3 \\
  Persp. & debate& 541/227 &sentence&  2 \\\bottomrule 
  \end{tabular}
  \caption{Dataset statistics.}\label{tab:data}
  \end{table}

\subsection{Datasets}
We choose datasets that can cover (i) multiple domains, (ii) different sizes of unseen topics, and (iii) various textual forms of topics (phrase-form and  sentence-form).
Therefore, we evaluate on three mainstream stance detection datasets:  SemT6 \cite{mohammad2016semeval}, VAST \cite{allaway2020zero} and  Perspectrum \cite{chen2019seeing}. We discard their training sets and dev sets to satisfy the definition of \taskname. 

\paragraph{SemT6 \cite{mohammad2016semeval}}  contains texts from the tweet domain regarding 6 topics: \texttt{Donald Trump},  \texttt{Atheism}, \texttt{Climate Change is a real Concern}, \texttt{Feminist Movement}, \texttt{Hillary Clinton}, and \texttt{Legalization of Abortion}. It is a three-way stance detection problem with labels \{\texttt{support}, \texttt{oppose}, \texttt{neutral}\}. Note that the prior applications of  SemT6 for zero-shot stance detection always trained on five topics and tested on the remaining one. To match the motivation of \taskname, we treat the whole SemT6 data as  \emph{test}, i.e., all six topics are unseen. When we report the data-specific supervised performance, we follow prior work to regard any five topics as seen and test on the sixth topic; each topic will have the chance to be unseen, and the average performance is reported.

\paragraph{VAST \cite{allaway2020zero}.}
In contrast to SemT6, VAST contains text from the New York Times ``Room for Debate'' section, and many more topics (4,003 in $train$, 383 in $dev$ and 600 in $test$). Those diverse topics, covering various themes, such as education, politics, and public health, are short phrases that are first automatically extracted and then modified by human annotators. Like SemT6, it also has three stance labels, but the \texttt{neutral} topics were randomly picked  from the whole topic set. For our \taskname~task, we only use its $test$ to evaluate our system and do not touch the gold labels of its $train$ and $dev$.

\paragraph{Perspectrum \cite{chen2019seeing}} is a binary stance detection benchmark (label is \texttt{support} or \texttt{oppose}) with two main distinctions with SemT6 and VAST: (i) both its text and topics were collected from several debating websites, and (ii) the topics are sentences rather than noun phrases.  Similar to VAST, we do not train our model on its $train$ and $dev$. The performance on $test$ will be reported. Since there are no neutral samples in this dataset, when the model is pretrained as a 3-way classifier, we set the probability threshold as 1/3 on the \texttt{oppose} label:  any prediction that has the \texttt{oppose} probabilities \emph{lower than 1/3} will be considered as \texttt{support}. Otherwise, the label would be \texttt{oppose}.

The detailed statistics of the three datasets are listed in Table \ref{tab:data}.

\begin{table*}[t]
 \setlength{\belowcaptionskip}{-5pt}
 \setlength{\abovecaptionskip}{5pt}
  \centering
  \begin{tabular}{llll|ccc|c}
  \toprule
  &&& & \multicolumn{4}{c}{F1 Score}\\
  &&& & SemT6 & VAST & Persp. & mean\\\midrule\midrule
  \multicolumn{4}{c|}{random guess} & 32.0 & 33.3 & 49.8 & 38.3\\
\multicolumn{4}{c|}{data-specific supervised learning (prior SOTA)}   &38.9 &78.0   &91.0  & 69.3 \\\midrule
\multicolumn{3}{l}{\multirow{3}{*}{cross-domain transfer}}  &SemT6 as \emph{train} &38.9 &28.9  &47.7 &38.5 \\\cmidrule(lr){4-8}
&&  &VAST as \emph{train} &55.4 &78.0 &49.0 &60.8 \\\cmidrule(lr){4-8}
&& &Pers as \emph{train}&26.7 &27.0  &91.0 &48.2 \\\hline
  \multirow{12}{*}{\rotatebox{90}{open-domain transfer}} & \multirow{3}{*}{\rotatebox{90}{baseline}} & & BERT &22.7 &36.8  &36.5 &32.0 \\
   &   & & GPT-3 &30.5 & 34.2  &39.9 &34.9 \\
   &   & & Cosine &31.5 &35.9  &62.7 &43.4 \\\cmidrule(lr){2-8}
    & \multirow{9}{*}{\rotatebox{90}{ours }} & \multirow{4}{*}{$D_{ind}$ and }& SemT6-based $D_{weak}$ &63.7 &69.8&82.8&72.1  \\
   &  &  & VAST-based $D_{weak}$ &64.3 &72.0  &80.4  &72.2\\
   &  &  & Persp-based $D_{weak}$ &64.5 &68.7  &79.5 &70.9 \\
   &  &  & joint $D_{weak}$ & 63.2  & 73.5 & 81.0 & \textbf{72.6} \\ \cmidrule(lr){3-8}
   & & & \enspace\enspace w/o indirect &49.6 &64.6  &38.2 & 50.8\\
   & & & \enspace\enspace w/o weak &45.3 &53.7  &79.1 &59.4 \\
      & & & \enspace\enspace w/o \masktopic &45.5 &65.2  &74.2 &61.6\\
   & & & \enspace\enspace w/o \masktext &63.4 &70.8 &78.2&70.8 \\\bottomrule 
  \end{tabular}
  \caption{Open-domain experiment results on  SemT6, VAST and Perspectrum. Our final number is in bold. }\label{tab:result}
  \end{table*}

\subsection{Baselines}

There are no prior systems that work on this new \taskname~problem since no training data is available. Here, we consider three baselines that can work on an unsupervised scheme.
\paragraph{BERT \cite{DBLPCLT19}.} Given the (\texttt{text}, \texttt{topic}) as input, ``BERT-large-uncased'' is used as a masked language model to predict the masked token in ``\texttt{text}, it [MASK] \texttt{topic}''. BERT will output the probabilities of the three label tokens \{\texttt{support}, \texttt{oppose}, \texttt{neutral}\} and the label that receives the highest probability would be the predicted stance.

\paragraph{GPT-3 \cite{brown2020language}.} Given the \texttt{text} and the \texttt{topic} with the instruction telling the model what task  we are trying to accomplish, GPT-3 is able to complete the prompt by choosing one of the given labels \{\texttt{support}, \texttt{oppose}, \texttt{neutral}\}. GPT-3 also has functions designed for classification, but the text completion scheme does a better job on this stance detection task. 
Our prompt:
\fbox{%
\small
  \begin{minipage}{0.95\linewidth}
     Given a topic and a text, determine whether the stance of the text is support, against, or neutral to the topic.\\
     \texttt{Topic}: Atheism\\
     \texttt{Text}: Everyone is able to believe in whatever they want. \\
     \texttt{Stance}: \rule{2cm}{0.25mm}
  \end{minipage}
}

\paragraph{Cosine similarity.} We compare the similarities between the \texttt{text} and a hypothesis sentence that combines \texttt{label} and \texttt{topic}, such as ``it \texttt{supports} the \texttt{topic}'', ``it \texttt{opposes} the \texttt{topic}'', or ``it is \texttt{unrelated} to the \texttt{topic}''. We first get the sentential representations by sentence-BERT \cite{DBLPeimersG19}, then choose the label whose resulting hypothesis obtains  the highest cosine similarity score.

In addition to the unsupervised baselines, we further consider the data-specific supervised training as the upperbound, and the following variants of our system: i) only \masktext~or \masktopic; ii) only indirect supervision or weak supervision.

\subsection{Setting}

\paragraph{GPT-3 for $D_{weak}$ collection.} The engine we chose for GPT-3 is ``curie'', which gives good quality at a reasonable price. There are several parameters that we played with. We set the temperature, which goes from 0 to 1 and controls the randomness of the completion generated, as 0.8 for \masktopic~and 0.9 for \masktext~for more diverse results. The randomness for \masktext is slightly higher because for some datasets the number of topics is extremely limited, such as SemT6, which only has 6 topics in total; therefore, we want to force diverse responses from GPT-3. The max number of tokens GPT-3 can generate is 6 for \masktopic~and 150 for \masktext. It is worth mentioning that GPT-3 will not necessarily generate as much as the upper bound, sometimes not even close. We let the stop word be "$\backslash$n", so that it stops generating when it reaches a new paragraph. ``top\_p''  is set as 1, letting all tokens in the vocabulary been used. ``frequency\_penalty'' is 0.3 for \masktext~to avoid the model producing the same line again and again. 

\paragraph{Training details.}
All models are optimized using AdamW \cite{DBLPhilovH19}. Learning rate  1e-6, batch size 16, maximal (premise, hypothesis) length is 200. The system is trained for 20 epochs on \emph{train} and the best model on \emph{dev} is kept. 



\subsection{Result}
Table \ref{tab:result} lists the main results. We first include ``data-specific supervised learning'' as the upperbound performance and the ``cross-domain transfer'' that takes each dataset as the source domain and tests on others respectively. Both settings try to explore the upper limit when we apply human-annotated supervision. Our core task, \taskname, is evaluated in the last three blocks. 

From the baseline block, we can  observe that for all domains, baseline methods mostly perform like random guess, except for the slight improvement of the ``cosine'' approach over Perspectrum. This result indicates the difficulty of the real-world \taskname~task we proposed. Although BERT and GPT-3 are the top-tier pre-trained language models, they still cannot handle \taskname~well. 

Then look at our approach that combines indirect supervision data ($D_{ind}$) and weak supervision data ($D_{weak}$). Note that $D_{weak}$ can be collected based on the  \emph{label-free train} of VAST, SemT6 or Perspectrum. We try $D_{weak}$ for each of the task domains and also put them jointly (i.e., ``joint $D_{weak}$''). We note that all four versions of $D_{weak}$ result in very consistent performance---mostly around 72\% by the ``mean''. This clearly supports the \emph{robustness of our method}: it is less affected by the original domain where \texttt{text} and \texttt{topic} come from, and a single system based on each of the domain or their combination can perform well on all domains. 

The last block of Table \ref{tab:result} reports the ablation study, where we discard individual source of supervision (indirect or weak) or individual masking scheme (\masktext~or \masktopic). We observe that i) indirect supervision and weak supervision play complementary roles for the task \taskname; and they both outperform baselines by large margins, and ii) both masking schemes help, and  the \masktopic~contributes more. This is maybe because \masktopic~requires the GPT-3 to generate shorter texts than \masktext~so that \masktopic~can yield higher-quality data. Additionally, deriving supporting sentences  for a given topic sometimes requires substantial background knowledge and solid reasoning, which is still a difficult task for GPT-3.




\subsection{Analysis}
Next, we conduct a deep analysis for the system robustness towards prompts ($\mathcal{Q}_1$), the required size of $D_{weak}$ ($\mathcal{Q}_2$), the noise in generated $D_{weak}$ ($\mathcal{Q}_3$), and the error patterns made by our system ($\mathcal{Q}_4$).
\paragraph{$\mathcal{Q}_1$: Robustness of dealing with prompts.} Prompt design takes place in both GPT-3 completion and the conversion from stance detection to textual entailment. 
When generating the prompt for GPT-3, how we construct the prompt in \masktopic~and \masktext~can make a huge impact on the completion received.  In \masktopic, we use the prompt ``He said  \texttt{text}, so he \texttt{label} the idea of [MASK]''. The reason why we add ``\emph{the idea of}'' at the end of the prompt is because it helps the model understand that we want a noun phrase. Otherwise, we will see completions like ``that'', ``it'', etc. Similarly, in \masktext, the final prompt we use is ``His attitude towards \texttt{topic} is \texttt{label} because he thinks [MASK]''. Considering the freedom of GPT-3 completion, we add ``s/he thinks'' at the end of the prompt, forcing GPT-3 to generate a reasoning for the given topic/label pair. If we don't add ``\emph{he thinks}'' at the end, it would be common to see GPT-3 repeating the given sentence in the generated completion. In addition, when the label is \texttt{neutral}, such as the prompt  ``His attitude towards high school writing skills is neutral because he thinks [MASK]'', GPT-3 would output sentences like ``he does not have a strong opinion either way'' if we don't have ``\emph{he thinks}'' at the end. After the modification, responses would make more sense, such as ``that they are important but not essential.'' These tricks in prompt design suggest that it is essential to make the sentence structure as clear as possible and provide content that helps to instruct the model on what we want.
\begin{figure}
 \setlength{\belowcaptionskip}{-10pt}
 \setlength{\abovecaptionskip}{5pt}
    \centering
    \includegraphics[width=1\linewidth]{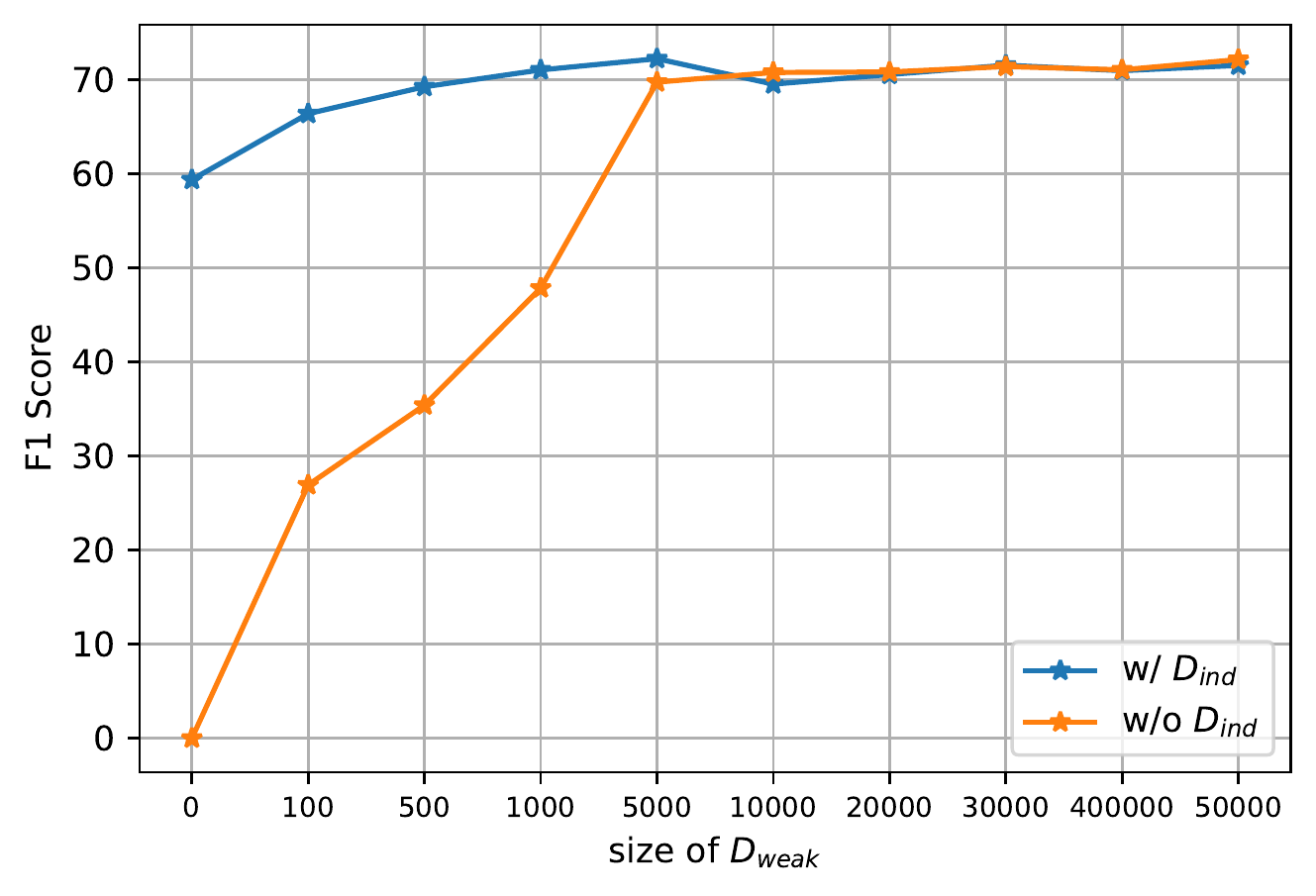}
    \caption{Mean F1  vs. size of $D_{weak}$.}
    \label{fig:weaksize}
\end{figure}

When we convert the topic phrase into a sentential hypothesis, we again get involved in the prompt design. During training, we stick with ``he is in favor of \texttt{topic}'' template to limit the training size, but in the testing, we found the majority voting of four templates (``he/she is in favor of \texttt{topic}'' and ``he/she opposes \texttt{topic}'') lead to comparable performance with ``he is in favor of \texttt{topic}''. This indicates the pre-trained entailment system is considerably robust in dealing with hypotheses derived from different templates.

\paragraph{$\mathcal{Q}_2$: How much weakly supervised data is needed?} We answer this question by applying $D_{weak}$ alone or together with $D_{ind}$. For each case, we test on  sizes varying from 100 to 50,000 and report the average results over 3 random seeds. From the Figure \ref{fig:weaksize}, we can see that both settings can reach similar performance when we collect over 10k data of $D_{weak}$, but the pretraining on $D_{ind}$ can dramatically reduce the required size of $D_{weak}$: from 10k to around 500.

\paragraph{$\mathcal{Q}_3$: Error patterns of weakly supervised data.} 
We collect typical error patterns in $D_{weak}$ derived by \masktopic~and \masktext~separately. 
\paragraph{\masktopic.} Three typical error types.

\textbullet \emph{Incomplete generation}. Sometimes GPT-3 fails to give a complete topic phrase and cuts in the middle even though it hasn't reached the maximum token limit. For example:
\fbox{%
\small
  \begin{minipage}{0.95\linewidth}
     He claims \texttt{16 year olds are informed enough to cast a vote}, so he \texttt{supports} the idea of \underline{\textsc{giving 16-year-olds}}
  \end{minipage}
}

In this example, the topic given by GPT-3 is ``giving 16-year-olds'', which is not a complete phrase as we expected. This kind of errors  indicate that GPT-3 sometimes stops generating before providing a complete idea even when the word limit is not exceeded.

\textbullet \emph{Failure in understanding the stance}. Since we are providing opposite labels (i.e., \texttt{support} and \texttt{oppose}), we hope that GPT-3 would  produce distinct topics that hold opposite stances. 
However, sometimes GPT-3 fails to understand the stances when generating topics. For example:
\fbox{%
\small
  \begin{minipage}{0.95\linewidth}
    He claims \texttt{A higher minimum wage means less crime}, so he \textcolor{blue}{\texttt{supports}} the idea of \underline{\textsc{a minimum wage}}\\
    He claims \texttt{A higher minimum wage means less crime}, so he \textcolor{red}{\texttt{opposes}} the idea of \underline{\textsc{a minimum wage}}
  \end{minipage}
}

This error type is the most common one in the weakly supervised data (approximately 85\% error instances), indicating that GPT-3 is still less effective to interpret negated information. 

\textbullet \emph{Misunderstanding the text}. The GPT-3 does not always understand the meaning of the sentence correctly. For example:
\fbox{%
\small
  \begin{minipage}{0.95\linewidth}
     He claims \texttt{women who are housewives should be paid}, so he \texttt{supports} the idea of \underline{\textsc{women being paid less than men}}
  \end{minipage}
}

Here, the predicted topic is related but not the main subject of the sentence. Such a mistake is  rare but still exists weak supervision.

\paragraph{\masktext.} 
Even though GPT-3 can mostly provide a sentence that is related to the topic and align with the correct stance, more than 50\% of the time the content is very short and less informative compared to the texts from the datasets. For example:
\fbox{%
\small
  \begin{minipage}{0.95\linewidth}
     His attitude towards $\texttt{middle east oil}$ is $\texttt{opposition}$ because he thinks \underline{$\textsc{it is a waste}$}
  \end{minipage}
}
\fbox{%
\small
  \begin{minipage}{0.95\linewidth}
     His attitude towards $\texttt{miss america}$ is $\texttt{support}$ because he thinks \underline{$\textsc{she is talented}$}
  \end{minipage}
}

This is not that surprising since GPT-3 was trained to mainly satisfy the language modeling criterion; thus, it would be ``lazy''  to return with a solid and long response. These \masktext~instances are never wrong in the judgment of attitudes, so they can still give the model some help, although limited, in determining the attitudes.


\paragraph{$\mathcal{Q}_4$: Error analysis of our system.} Due to space limitation, we summarize two common error patterns made by our system.

\textbullet \emph{Failed to connect the topic and text}. The text often mentions the topic with distinct expressions and contains its stance implicitly. Therefore, it brings more difficulty to the model to successfully locate the topic and identify the stance. For example:
\fbox{%
\small
  \begin{minipage}{0.95\linewidth}
     \texttt{Topic}: musician  \\
     \texttt{Text}: Spotify and Pandora pay usage rates that are much lower than the radio, records and legal downloads that they are replacing. Low enough to where many potential new artists won't be able to even earn a living. There must be some alternative other than artists simply being forced to accept the new streaming model that destroys royalties. For example, who set streaming royalty rates? Can artists unionize and negotiate collectively with the streaming services? If we don't sort this out, we will lose a new generation of artists -- which is bad for everyone.\\
     Gold \texttt{label}: support\\
     Predicted \texttt{label}: neutral
  \end{minipage}
}

\textbullet \emph{Incorrect ground-truth labels.}
The gold labels are not always correct. Sometimes the model makes a more appropriate judgement than the data provides. For example:
\fbox{%
\small
  \begin{minipage}{0.95\linewidth}
     \texttt{Topic}: keep weight \\
     \texttt{Text}: ``All the medical evidence points to the fact that it's nearly impossible to keep off weight once lost. The body just won't let you." This is incorrect, and could lead to fatalism that could harm people who are overweight. For example, I lost 70 pounds. That was at least a year ago. It has not come back. It is easy to keep off......'' \\
     Gold \texttt{label}: neutral \\
     Predicted \texttt{label}: support 
  \end{minipage}
}


\section{Conclusion}
In this work, we define \taskname, a more realistic and challenging zero-shot stance detection problem in an open world. Under such a setting, multiple domains and numerous topics can be involved, while no topic-specific annotations are required. To solve this problem, we proposed to combine   indirect supervision from textual entailment and weak supervision collected  from GPT-3. Our system, without the help of any task-specific supervision, outperforms the supervised method on three benchmark datasets that cover various domains and free-form topics. 

\section*{Acknowledgment}

The authors appreciate the reviewers  for their insightful comments and suggestions.

\bibliography{anthology}
\bibliographystyle{acl_natbib}




\end{document}